\documentclass[sigconf]{acmart}

\renewcommand\footnotetextcopyrightpermission[1]{} 
\usepackage{algorithmic}
\usepackage{graphicx}
\usepackage{textcomp}
\usepackage{xcolor}
\usepackage{url}
\usepackage{graphicx}
\usepackage{multirow}
\usepackage{booktabs}
\usepackage{graphics}
\usepackage{comment}
\usepackage{dirtytalk}
\usepackage{multirow}
\usepackage{tablefootnote}
\usepackage[normalem]{ulem}

\newcommand{\isabel}[1]{\textcolor{green!80!black}{\textbf{[ID: #1]}}}
\newcommand{\lu}[1]{\textcolor{blue!80!black}{\textbf{[LC: #1]}}}

\begin{document}

\acmConference{Preprint}{July 2022}{arXiv}

\title{Towards a Sentiment-Aware Conversational Agent}

\author{Isabel Dias}
\email{isabel.h.dias@tecnico.ulisboa.pt}
\affiliation{%
  \institution{INESC-ID, Instituto Superior T\'{e}cnico, Universidade de Lisboa}
  \country{Portugal}}

\author{Ricardo Rei}
\email{ricardo.rei@unbabel.com}
\affiliation{%
  \institution{Unbabel, INESC-ID, Instituto Superior T\'{e}cnico, Universidade de Lisboa}
  \country{Portugal}}
  
\author{Patr{\'\i}cia Pereira}
\email{patriciaspereira@tecnico.ulisboa.pt}
\affiliation{%
  \institution{INESC-ID, Instituto Superior T\'{e}cnico, Universidade de Lisboa}
  \country{Portugal}}

\author{Luisa Coheur}
\email{luisa.coheur@tecnico.ulisboa.pt}
\affiliation{%
  \institution{INESC-ID, Instituto Superior T\'{e}cnico, Universidade de Lisboa}
  \country{Portugal}}

\begin{abstract}
In this paper, we propose an end-to-end sentiment-aware conversational agent based on two models: a reply sentiment prediction model, which leverages the context of the dialogue to predict an appropriate sentiment for the agent to express in its reply; and a text generation model, which is conditioned on the predicted sentiment and the context of the dialogue, to produce a reply that is both context and sentiment appropriate. Additionally, we propose to use a sentiment classification model to evaluate the sentiment expressed by the agent during the development of the model. This allows us to evaluate the agent in an automatic way. Both automatic and human evaluation results show that explicitly guiding the text generation model with a pre-defined set of sentences leads to clear improvements, both regarding the expressed sentiment and the quality of the generated text.
\end{abstract}


\maketitle

\section{Introduction}\label{section:introduction}

End-to-end data-driven conversational agents have become popular over the last years. Current works focus on ways to explicitly condition text generation models in order to produce certain attributes, such as emotions \cite{zhou2018emotional, huang2018automatic}. 

\begin{figure}[!ht]
    \centering
    \includegraphics[width=0.8\columnwidth]{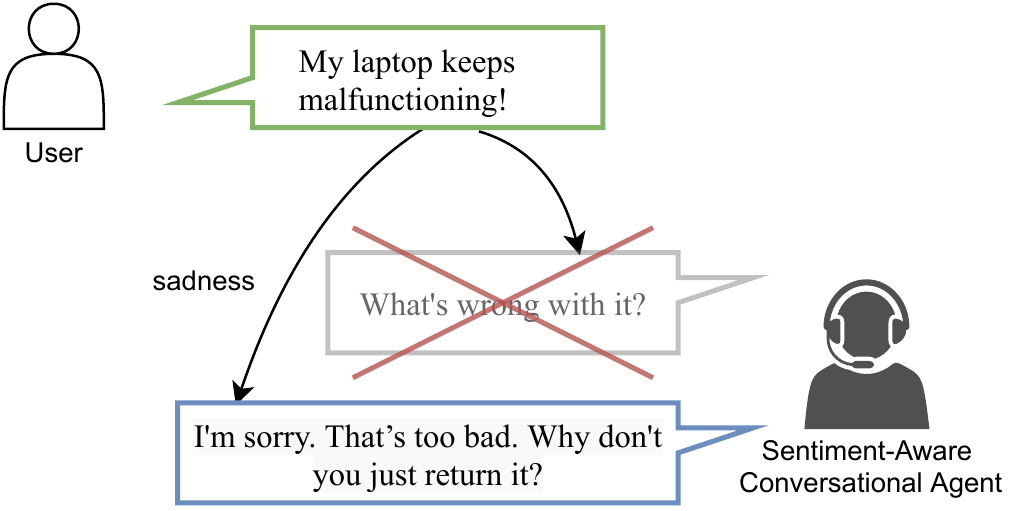}
    \caption{Dialogue generated by Baseline vs. SACA conditioned on \textit{Sadness} sentiment.}
    \label{fig:thesis-example}
    \vspace{-4mm}
\end{figure}

In fact, generating answers with the appropriate emotion can be particularly important in scenarios as, for instance, customer support. In the example of Figure \ref{fig:thesis-example}, while both answers can be considered correct, a reply that expresses an appropriate emotion may be more satisfying for the user, rather than the generic reply.

As highlighted in \cite{li2018emotion}, in human-human conversations the emotions expressed in two subsequent utterances from different speakers often change and are important when predicting the \say{correct emotion for an upcoming response before generation}. According to \cite{xie2021generating}, the drawback of current approaches is that they can not be used in a dialogue setting, given that they do not introduce a mechanism to automatically predict the next appropriate attribute.  

\begin{figure}[ht!]
    \centering
    \includegraphics[width=\columnwidth]{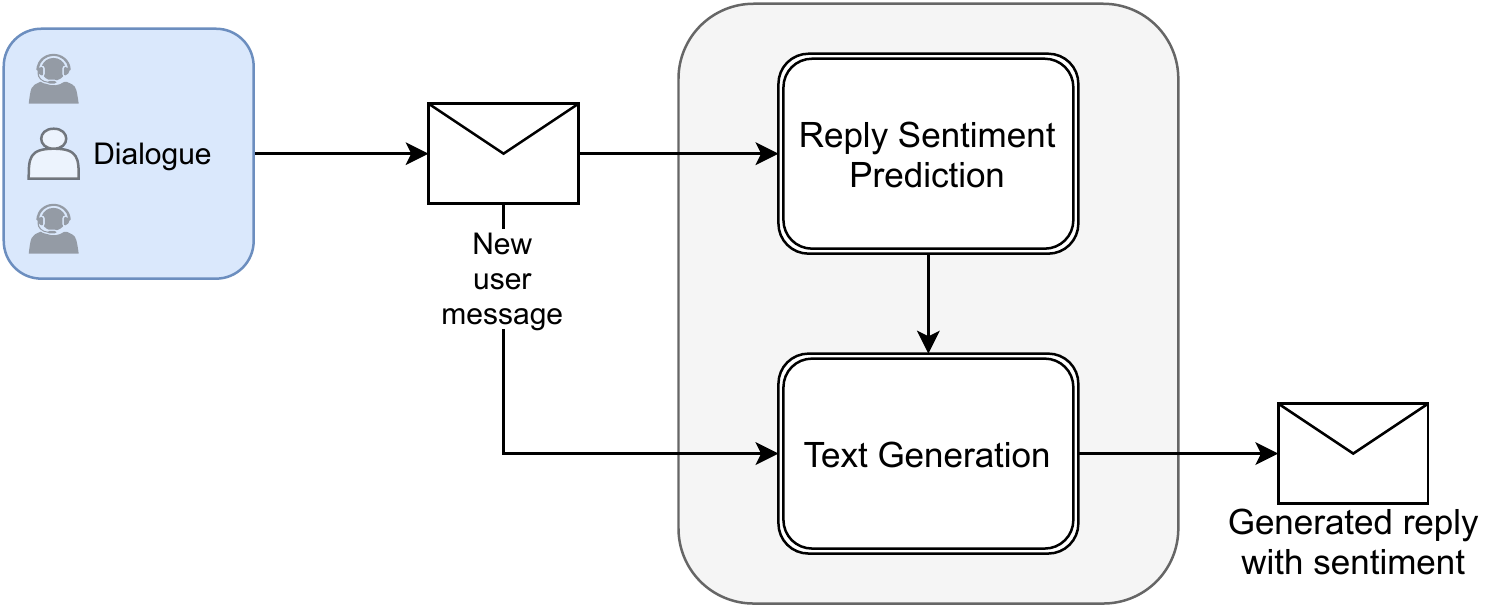}
    \caption{Proposed sentiment-aware conversational agent.}
    \label{fig:thesis-diagram}
    \vspace{-2mm}
\end{figure}

In this paper, we propose an \textbf{end-to-end sentiment-aware conversational agent} that predicts the next appropriate attribute and generates its answers accordingly. Figure \ref{fig:thesis-diagram} depicts the proposed architecture, which is based on two main models: 
\begin{itemize}
    \item A \textbf{reply sentiment prediction model}, which predicts the appropriate reply sentiment that should be expressed by the conversational agent in the next utterance;
\item A \textbf{text generation model}, which generates a sentence that is context-aware and expresses the predicted sentiment.
\end{itemize}

We also propose to use a \textbf{sentiment classifier} to evaluate the generated sentences. This allows us not only to understand whether the model is expressing the correct sentiment, but also to do so in an automatic manner during the development of the model. Furthermore, we explore \textbf{retrieval augmentation} in order to bias the classification models towards the correct sentiment label.

\begin{table*}[!ht]
\small
\centering
\begin{tabular}{ll|ll}
\toprule
\textbf{Sentence} & \textbf{Label} & \textbf{Representation} & \textbf{Label} \\
\hline
Does it cost anything? & \textsc{Neu} & \texttt{[CLS]Does it cost anything?[SEP]} & \textcolor{green}{\textsc{Neu}} \\
Yeah 20\$ per month. & \textcolor{green}{\textsc{Neu}} & \texttt{[CLS]Yeah 20\$ per month.[SEP]Does it cost anything?[SEP]} & \textcolor{blue}{\textsc{Sur}} \\
Ohh! & \textcolor{blue}{\textsc{Sur}} & - & -  \\
\toprule
\end{tabular}
\caption{Example of a dialogue from the EmotionPush dataset along with the sentence representation and corresponding label for the reply sentiment prediction task. \textsc{Neu} corresponds to the label \textit{Neutral}, and \textsc{Sur} to the label \textit{Surprise}.}
\label{tab:methods-contextual-nsp-example}
\vspace{-6mm}
\end{table*}

In addition, we study the best way to add information regarding the appropriate sentiment, in order to guide the text generation model. To this end, several \textbf{sentiment's lexicons}, such as simple sentiment tags or sentences capturing the different sentiments are considered; as we will see, feeding the model with a small set of sentences expressing emotions leads to the best results. In fact, our results show that: a) using the appropriate sentiment, improves, indeed, the scores in the generation process; b) that the evaluation based on the sentiment classification model correlates well with the human evaluation performed; c) although following current state of the art architectures, there is still plenty of room for improvement in the  reply sentiment prediction model; and that d) the retrieval augmentation process did not significantly improve results. 

We test our models with the EmotionPush \cite{hsu-etal-2018-emotionlines} and DailyDialog \cite{li2017dailydialog} corpora, which use as labels the six Ekman's basic emotions \cite{ekman1992argument}, and neutral.

\section{The Sentiment Aware Agent}
\label{chap:sentiment-analysis}

In this section, we describe the proposed sentiment classification model, reply sentiment prediction model, and text generation model, conditioned on the predicted sentiment and on the past conversation context. We also describe the retrieval augmentation techniques used.

\subsection{The Sentiment Classification and Reply Sentiment Prediction Models}

For both sentiment classification and reply sentiment prediction we used pre-trained Transformer models, such as BERT \cite{devlin2019bert}. We improved upon this base model, by using previous dialogue context. In order to add context to the input of the model, we took advantage of a particularity of the BERT architecture, the \texttt{[SEP]} token, by considering that the first sentence after the \texttt{[CLS]} token is the sentence we are trying to classify, and it is followed by its context. Thus, given a dialogue $D = (s_{1}, s_{2}, ...,  s_{n})$, with $n$ equal to the number of sentences in the dialogue, in order to classify the sentence $s_{i}$ with $x$ sentences as context, the input to the model is $\mathcal{}{concat}(s_{i}, s_{i-1}, ..., s_{i-x})$. 

Regarding the reply sentiment prediction model, similarly to the contextual sentiment classification, we also made use of the \texttt{[SEP]} token to separate the different sentences that are part of the input. However, there are two main differences: first, the input of the reply sentiment prediction model corresponds only to the previous context sentences; secondly, the gold label is the sentiment of the upcoming sentence. 
In this task we can also use an arbitrary number of sentences as context. We will use as default setup the last two previous subsequent sentences. In Table \ref{tab:methods-contextual-nsp-example} we can observe an example of the input of the model.

\subsection{The Text Generation Model}
\label{section:text-generation-model}

The text generation model receives the sentiment predicted by the reply sentiment prediction model, as well as the previous context of the conversation, to generate a sentence that is not only appropriate given a context, but also expresses the predicted sentiment. To do so, we adapted part of the work in \cite{wolf2019transfertransfo}, that proposes a model that is able to leverage a set of sentences that describe a given persona in order to generate text that is coherent with that persona. We adapted this concept and developed a \textbf{sentiment lexicon knowledge base}, which was used to make the conversational agent sentiment-aware.  In order to guide the model towards an appropriate sentiment, we concatenated the desired sentiment's lexicon to the input of the model. The sentiment lexicon for each sentiment was built using the following techniques: 
\begin{itemize}
    \item \textbf{Tag}: use the sentiment's name (e.g., anger, joy, etc.). We will use this model as a baseline for the sentiment conditioned text generation model;
    \item \textbf{TF}: retrieve the 40 most frequent $n$-grams from the set of training sentences of each sentiment. We assume $1 \le n \le 3$;
    \item \textbf{TFU}: the same as the \textbf{TF} approach, but removing $n$-grams that are not unique to each sentiment. We assume $1 \le n \le 3$;
    \item \textbf{TF-IDF}: retrieve 40 $n$-grams with the highest TF-IDF score. We assume $1 \le n \le 3$;
    \item \textbf{Random Sample}: select at random a sentence from the training set labelled with the sentiment we want to generate;
    \item \textbf{Sentiment Sentences}: use a pre-defined set of sentences representative of each sentiment. The set we are going to use can be observed in Table \ref{tab:tg-sentiment-sentences}. This set of short sentences was created using always the same sentence structure (\say{I am...} and \say{That is...}). For the DailyDialog corpus we do not include the \textit{non-neutral} label.
\end{itemize}

\begin{table}[ht!]
\centering
\begin{tabular}{c|ll}
\toprule
\textbf{Sentiment}              & \textbf{Sentence 1} & \textbf{Sentence 2} \\ \hline
\textbf{Anger} & I am angry. & That is so annoying! \\
                       
\textbf{Disgust} & I am disgusted. & That is repulsive! \\
                       
\textbf{Fear} & I am frightened. & That is scary! \\ 
                       
\textbf{Joy} & I am happy. & That is delightful! \\
                       
\textbf{Neutral} & I am ok. & That is ok. \\ 
                       
\textbf{Non-neutral} & I am not ok. & That is not ok. \\
                       
\textbf{Sadness} & I am sad. & That is so upsetting. \\
                       
\textbf{Surprise} & I am surprised. & That is so amazing! \\  
\toprule                       
\end{tabular}
\caption{Sentences used to represent each sentiment.}
\label{tab:tg-sentiment-sentences}
\end{table}

\subsection{The Retrieval Augmentation Model}

Due to the recent success of retrieval augmentation approaches in NLP tasks \cite{dinan2018wizard, moghe2018towards, lewis2020retrieval}, including sentiment classification \cite{ramos2021retrieval}, we also explored a mechanism that relies on nearest neighbors. Following \cite{ramos2021retrieval}, the first step was to find the nearest training example for each train/development/test example, and the corresponding label. To do so, we used the Sentence Transformer \cite{reimers-2019-sentence-bert}\footnote{\url{https://www.sbert.net}} library to create sentence embedding representations of all examples using the \texttt{paraphrase-distilroberta-ba\-se-v1} model. Next, we used the FAISS \cite{johnson2019billion}\footnote{\url{https://faiss.ai}} library to build an index with the sentence embeddings that belong to the train set in order to find the closest training examples. In particular, we chose the Euclidean distance to calculate the distances between the examples. After retrieving the nearest neighbors information, the second step is to incorporate it in the Transformer model. We started by initializing an extra set of embeddings, which we will call Sentiment Embeddings ($\mathrm{SE}$), one for each sentiment label, that are trained along with the model. We initialized the embedding of each sentiment with the average of the sentences' corresponding to a given sentiment. E.g., the sentiment embedding for \textit{joy} was initialized with the average of the embeddings corresponding to the sentences labelled with the sentiment \textit{joy} in the training set. Then, for each example, we incorporated the nearest training example label in the Transformer model, by concatenating the corresponding $\mathrm{SE}$ to its output, after pooling, which will be the input to the classification layer. 
\section{Experimental Setup}
\label{section:experiments}
This section describes the datasets, evaluation metrics, and setups used during this work\footnote{The code is publicly available here: The repository will be made available.}.

\subsection{Datasets}
We tested our models in the EmotionPush \cite{hsu-etal-2018-emotionlines} and DailyDialog \cite{li2017dailydialog} corpora. The former is composed of 1000 private conversations from Facebook Messenger. The latter, with dialogues in daily life scenarios, contains 13118 multi-turn dialogues, divided in 10 themes such as: Finance, Politics, Health, Work, etc. Both corpora use as labels the six Ekman's basic emotions \cite{ekman1992argument}, and neutral. The EmotionPush dataset has an additional label, non-neutral, to represent sentences in which the annotators did not reach consensus. 

\subsection{Evaluation Metrics}

We evaluated the sentiment classification and the reply sentiment prediction models using the micro (m) and macro (M) F1 averages. Given that both corpora are highly unbalanced, with the \textit{neutral} label composing over 80\% of the examples, we also evaluated our models both with the majority class and without the majority class (micro/Macro-No Majority Class metric which we refer to as m/M-NMC-F1). This allowed us to have a better understanding of how the models are performing on less represented sentiments.

Regarding the text generation model, our choice of automatic evaluation metrics aims to measure if the model is able to generate a sentence that expresses a desired sentiment without compromising the quality of the text. Therefore, first, regarding the quality of the generated text, we focused on Perplexity (\textit{PPL}) \cite{10.5555/1214993}, and Sentence Embedding Similarity (\textit{SES}) \cite{xie2021generating}. Second, to evaluate if the generated text is expressing the appropriate sentiment, we used the sentiment classification model to classify the generated sentences and evaluate them using the aforementioned F1 metrics. 

\subsection{Training Setup}

The models were implemented using PyTorch Lightning \cite{Falcon_PyTorch_Lightning_2019} and the HuggingFace Transformers \cite{Wolf_Transformers_StateoftheArt_Natural_2020} library.

The sentiment classification and reply sentiment prediction models were trained for a maximum of 40 epochs, using the cross entropy loss, with 4 validation steps per epoch, stopping after 10 consecutive validation steps without improvement. The checkpoint used to evaluate the model was the one that achieved the highest validation macro-F1 value. We follow the work by \cite{howard2018universal} and use the Adam optimizer \cite{kingma2014adam} with a discriminative learning rate of $1 \times 10^{-3}$, except for the Transformer model that has a learning rate of $5 \times 10^{-6}$. For the Transformer model we apply a  layer-wise learning rate decay of $0.95$ after each step. We apply a dropout \cite{srivastava2014dropout} of $0.4$ to the sentence embeddings during training. We use a real batch size of 16 whenever the GPU's memory allowed it, but used gradient accumulation to always simulate a batch size of 32.

The text generation models were trained for a maximum of 40 epochs, using the negative log-likelihood loss, with 4 validation steps per epoch, stopping the training after 12 consecutive validation steps without improvement. We use the Adam optimizer \cite{kingma2014adam} with a learning rate of $5 \times 10^{-6}$. The checkpoint used to evaluate the model was the one that achieved the lowest validation negative log-likelihood loss value. Due to computational constraints, we always use the two most recent context sentences from the dialogue as input to the model. We use a real batch size of 4 whenever the GPU's memory allowed it, but we used gradient accumulation to always simulate a batch size of 16.  All other hyperparameters were kept as default.

\subsection{Models Setup}
\label{section:models-setup}

Considering sentiment classification, and by using the development set, we empirically found the following optimal setup: a RoBERTa-large model \cite{liu2019roberta}, that receives as input the concatenation of the sentence to be classified, with the last previous context sentence; a linear classification layer that receives as input the concatenation of the \texttt{[CLS]} token embedding of the last 4 hidden layers (\textit{concat4} pooling); and uses the retrieval augmentation method previously described. As a baseline we considered a RoBERTa-large model with a linear classification layer, that receives as input the sentence to be classified.

Regarding the reply sentiment prediction task, and by also using the development set, the optimal setup found was: a RoBERTa-large model, that receives as input a concatenation of the last four context sentences; and a linear classification layer. The Retrieval augmentation did not improve performance and therefore was not used for this task. As a baseline we considered a RoBERTa-large model with a linear classification layer, that receives as input the last two context sentences.

Finally, our conditioned text generation model consists on a DialoGPT-small model, with the concatenation of the desired sentiment's lexicon to the input of the model.

\section{Results}

\subsection{Sentiment Classification}

We start by presenting the results of our Sentiment Classification model (from now on ``SentClass'' Model) (Table \ref{tab:sc-test-emotionpush}).

\begin{table}[ht!]
\small
\centering
\begin{tabular}{l|cccc}
\toprule
\multicolumn{5}{c}{\textbf{EmotionPush}} \\
\toprule
\textbf{Model} & \textbf{m-F1} & \textbf{m-NMC-F1} & \textbf{M-F1} & \textbf{M-NMC-F1} \\ \hline
Baseline & \textbf{78.9} & 57.6 & 45.4 & 39.2 \\
SentClass & \textbf{78.9} & \textbf{58.2} & \textbf{54.1} & \textbf{49.2} \\ \hline
+ RoBERTa-base & -0.7 & -0.8 & -4.2 & -4.7 \\
+ CLS & -0.9 & -0.9 & -1.6 & -1.7 \\
- Context 1 & -2 & -2.6 & -8.8 & -9.9 \\
- Ret. Aug. & -0.7 & +0.3 & +3.1 & +3.6 \\
\toprule
 \multicolumn{5}{c}{\textbf{DailyDialog}} \\
 \toprule
\textbf{Model} & \textbf{m-F1} & \textbf{m-NMC-F1} & \textbf{M-F1} & \textbf{M-NMC-F1} \\ \hline
Baseline & 84.5 & 56.5 & 48.1 & 41.0 \\
SentClass & \textbf{85.0} & \textbf{58.1} & \textbf{51.0} & \textbf{44.4} \\ \hline
- Context 1 & -0.6 & -1.6 & -2.5 & -3.0 \\
- Ret. Aug. & -0.6 & -1.3 & -1.0 & -1.2 \\ \hline
KET \cite{zhong2019knowledge} & - & 53.4 & - & - \\
ELECTRA\tablefootnote{with Contextual Augmentation} \cite{kim2020contextual} & - & 57.9 & - & - \\
COSMIC \cite{ghosal2020cosmic} & - & 58.5 & 51.1 & - \\
\toprule
\end{tabular}
\caption{Baseline vs. SentClass Model + Ablation Study.}
\label{tab:sc-test-emotionpush}
\vspace{-6mm}
\end{table}

Regarding the experiments done on the EmotionPush dataset, when compared with the baseline, our model improves all metrics, except the m-F1, where it maintains the same value. More notably, it is able to improve the M-F1 metric by 8.7 points. These improvements are also noticeable on the M-NMC-F1 metric, where our model improves 10 points.

In order to further validate our results we performed an ablation study using the test set, also displayed in Table \ref{tab:sc-test-emotionpush}, with four experiments defined as follows: \textbf{+RoBERTa-base}, where we replace RoBERTa-large by a RoBERTa-base; \textbf{+CLS}, where we replace the \textit{concat4} pooling by the embedding of the \texttt{[CLS]} token of the last hidden layer; \textbf{-Context 1}, where we no longer use context in the input of our model; \textbf{-Ret. Aug.}, where we remove the retrieval augmentation methods from the model. Results showed that removing the context is what impacts the model the most, which tells us it was the most significant addition to our model. Additionally, replacing the RoBERTa-large by the RoBERTa-base and the \textit{concat4} by the \textit{CLS} pooling option, also worsens all metrics. Interestingly, in the test set, using retrieval augmentation worsened our results. 

Regarding the results on the DailyDialog corpus, since on the ablation study performed on the EmotionPush corpus we found that removing context and retrieval augmentation had the most impact on the model, we focused only on those changes for this corpus. We can observe that all metrics improve significantly when compared to the baseline. In particular, the m-F1 metric improves by 0.5 points, and without the majority class by 1.6 points. Regarding the M-F1 metric, it improves by 2.9 points and without the majority class by 3.4 points. Both removing the context and retrieval augmentation worsens the results. 
We should also notice that our model outperforms all approaches except when compared with the work from \cite{ghosal2020cosmic} that makes use of an additional large knowledge base that captures certain aspects such as personality, or emotion interactions, although the performance is quite similar\footnote{For EmotionPush we can only compare our results with the ones from \cite{khosla2018emotionx}, although the metric used is the Unweighted Accuracy. We also outperform this work: 62.2 vs. 70.7 (ours). All the other works that use this dataset provide different splits.}.


\subsection{Reply Sentiment Prediction}
\label{subsection:nsp-exps-contextual}

In Table \ref{tab:comp-models-nsp-emotionpush} we can observe the comparison of our reply sentiment prediction (RSP) model. As described in Section \ref{section:models-setup}, we use a RoBERTa-large which receives as input a concatenation of the last four context sentences. As a baseline, we considered the same model, but receiving as input the last two context sentences.

\begin{table}[ht!]
\small
\centering
\begin{tabular}{l|cccc}
\toprule
\multicolumn{5}{c}{\textbf{EmotionPush}} \\
\toprule
\textbf{Model} & \textbf{m-F1} & \textbf{m-NMC-F1} & \textbf{M-F1} & \textbf{M-NMC-F1} \\ \hline
Baseline & \textbf{69.0} & 18.0 & 15.0 & 5.5 \\
RSP Model & 66.5 & \textbf{19.6} & \textbf{17.8} & \textbf{8.8} \\ \hline
+ RoBERTa base & +3.4 & -6.8 & +2.5 & -3.1 \\
+ CLS & +2.6 & -4.9 & -1.6 & -2.1  \\
+ Context 2 & +0.2 & -3.4 & -1.3 & -1.5 \\
\toprule
\multicolumn{5}{c}{\textbf{DailyDialog}} \\
\toprule
\textbf{Model} & \textbf{m-F1} & \textbf{m-NMC-F1} & \textbf{M-F1} & \textbf{M-NMC-F1} \\ \hline
Baseline & \textbf{80.7} & 40.1 & 33.8 & 24.6 \\
RSP Model & 80.4 & \textbf{42.8} & \textbf{35.0} & \textbf{26.1} \\ \hline
+ RoBERTa base & -1.2 & -1.5 & -2.9 & -3.2 \\
+ CLS & +0.7 & -1.3 & -0.4 & -0.6  \\
+ Context 2 & -0.2 & -1.7 & +0.5 & +0.5 \\
\toprule
\end{tabular}
\caption{Baseline vs. RSP Model + Ablation Study.}
\label{tab:comp-models-nsp-emotionpush}
\vspace{-4mm}
\end{table}

We can start by observing that only the m-F1 does not improve when compared to the baseline. Nonetheless, our model outperforms the baseline in all other metrics (m-NMC-F1 by 6.7 points, M-F1 by 2.4 points, and M-NMC-F1 by 2.9 points). This shows that our model is better at generalizing for less represented sentiments.

In order to further validate our results, we performed another ablation study, which can also be observed in Table \ref{tab:comp-models-nsp-emotionpush}. In this study, all models have a higher m-F1 than the RSP model. On the remaining metrics all perform worse than the RSP model. These results show how our model is doing a trade-off between a lower F1 in the majority class and a higher F1 on the less represented sentiments.

Regarding the DailyDialog corpus, the results obtained show that our introduced changes also improve the baseline. Furthermore, we can observe that even using a larger number of training examples, the reply sentiment prediction task is still hard to perform well in. Finally, regarding the ablation study, we can observe that all introduced changes result in an improvement on most metrics.  


\subsection{Selecting the Best Sentiment Lexicon}
In order to understand which sentiment lexicon resulted in the best performance, we applied each lexicon to a Dialo-GPT-small text generation model, concatenating the appropriate sentiment lexicon to the input of the model, as described in Section \ref{section:text-generation-model}. This experiment was done using the EmotionPush dataset. The results can be observed in Table \ref{tab:tg-sentiment}. 

\begin{table}[!ht]
\small
\centering
\begin{tabular}{l|cc|cccc}
\toprule
 & \multicolumn{2}{c|}{} & \multicolumn{4}{c}{\textbf{F1}} \\
\textbf{Sentiment Lexicon} & \textbf{\textsc{PPL}} & \textbf{\textsc{SES}} & \textbf{m} & \textbf{m-NMC} & \textbf{M} & \textbf{M-NMC}  \\ \hline
None & 92.4 & 16.8 & 42.5 & 15.6 & 13.2 & 6.4 \\
Tag & 90.1 & 17.7 & 45.6 & 16.8 & 14.2 & 7.2 \\
TF & 91.2 & 16.4 & 45.5 & 14.4 & 12.6 & 5.6 \\
TFU & 90.7 & 16.3 & 45.0 & 14.7 & 13.8 & 6.7 \\
TF-IDF & 120.9 & 16.4 & 41.7 & 13.6 & 12.6 & 5.7 \\
Random Sample & 94.3 & 16.8 & 44.7 & 20.1 & 15.0 & 8.3 \\
Sentiment Sentences & \textbf{86.3} & \textbf{18.0} & \textbf{62.7} & \textbf{42.3} & \textbf{29.1} & \textbf{22.4} \\
\toprule
\end{tabular}
\caption{Results of concatenating sentiment lexicon to the input of the DialoGPT-small text generation model using the EmotionPush dataset.}
\label{tab:tg-sentiment}
\end{table}

Regarding the perplexity, we can observe that all approaches, except the \textit{TF-IDF} and the \textit{Sentiment Sentences}, perform similarly to the baseline (\textit{None}). When compared to the baseline, the \textit{TF-IDF} approach worsens the perplexity by 28.5 points, while the \textit{Sentiment Sentences} approach improves this metric by 6.1. Furthermore, the \textit{Sentiment Sentences} approach also improves the SES metric by 1.2 points. More interestingly, this approach improves significantly the metrics that are evaluating the expressed sentiment. When compared to the baseline, the m metric improves by 20.2 points, the m-NMC by 26.7 points, the M by 15.9 points, and the M-NMC by 16 points. These results show how this model is not only generating better sentences, due to the improvements on the perplexity and the SES, but also expressing the correct sentiment more times, which can be concluded by the improvements on the metrics that evaluate the expressed sentiment.

\subsection{Text Generation}
\label{section:exp-text-generation}
In Table \ref{tab:tg-emotionpush}, we can observe the results obtained in the text generation experiments. For these, we consider two baselines: the first (\textit{BL}) is a DialoGPT-small not conditioned on sentiment; and the second (\textit{Tag}) is a DialoGPT-small conditioned on the sentiment tags. Additionally, we show the results obtained using our best performing model, (\textit{SM}) a DialoGPT-small conditioned with the pre-defined set of sentiment sentences defined in Table \ref{tab:tg-sentiment-sentences}.

Both \textit{Tag} and \textit{SM} models are able to improve the perplexity. Furthermore, these models perform exceptionally well when compared to the \textit{BL} model on the sentiment-related metrics, increasing all metrics significantly. Nonetheless, the \textit{SM} approach yields the best overall results.

\begin{table}[!ht]
\small
\centering
\begin{tabular}{l|cc|cccc}
\toprule
\multicolumn{7}{c}{\textbf{Emotion-Push}} \\
\toprule
& \textbf{\textsc{PPL}} & \textbf{\textsc{SES}} & \textbf{m-F1} & \textbf{m-NMC-F1} & \textbf{M-F1} & \textbf{M-NMC-F1}  \\ \hline
BL & 85.0 & 17.3 & 47.6 & 22.6 & 17.1 & 10.3 \\
Tag & \textbf{78.9} & 16.7 & 61.4 & 43.9 & 24.2 & 17.4 \\
SM & 79.4 & \textbf{18.0} & \textbf{64.3} & \textbf{45.5} & \textbf{33.0} & \textbf{26.6} \\
\toprule
\multicolumn{7}{c}{\textbf{DailyDialog}} \\
\toprule
  & \textbf{\textsc{PPL}} & \textbf{\textsc{SES}} & \textbf{m-F1} & \textbf{m-NMC-F1} & \textbf{M-F1} & \textbf{M-NMC-F1}  \\ \hline
 BL & 9.9 & 26.7 & 77.9 & 30.6 & 24.9 & 14.5 \\
 Tag & \textbf{9.5} & \textbf{28.5} & 83.6 & 51.2 & 42.6 & 34.6 \\
 SM & 9.6 & 27.3 & \textbf{84.6} & \textbf{53.1} & \textbf{48.5} & \textbf{41.4} \\
\toprule
\end{tabular}
\caption{Results with the three different models.}
\label{tab:tg-emotionpush}
\vspace{-2mm}
\end{table}

Contrarily to the results observed for the EmotionPush corpus, the results obtained on the DailyDialog corpus, show that the \textit{Tag} and the \textit{SM} models perform more similarly. However, it is relevant to mention that the \textit{SM} model also outperforms the \textit{Tag} model on the sentiment metrics, more significantly on the macro metrics. 

As previously mentioned, the EmotionPush corpus is retrieved in an online chat context, which means the text is very informal, while the DailyDialog corpus was built from websites that are used to practice English, which makes the corpus more formal and fluent. This aspect could influence the quality of the generation models. In particular, the fact that the \textit{Tag} and \textit{SM} models fine-tuned with the DailyDialog corpus perform similarly could be an indication that the quality of the data used makes the models fine-tuned for this corpus not as dependent on the provided set of sentences, and a more simple option, such as a sentiment tag, is enough to guide the models. In contrast, the informality of the EmotionPush corpus could be making the models fine-tuned for this dataset more reliant on full sentiment sentences in order to generate good quality text conditioned on a sentiment.


\subsection{Sentiment-Aware Conversational Agent}

In order to fully evaluate the sentiment-aware conversational agent, we considered three models: 
\begin{itemize}
    \item the \textit{Baseline}, the DialoGPT-small model, which is not conditioned on sentiment;
    \item the \textit{Oracle}, the DialoGPT-small + pre-defined set of sentences model. Since this model is conditioned on the gold sentiment label, it represents the proposed sentiment-aware conversational agent if the RSP model was perfect;
    \item \textit{SACA}, the proposed Sentiment-Aware Conversation Agent. This agent consists on the best performing models obtained empirically. The RSP model (RoBERTa-large + four context sentences), and the text generation model conditioned on the RSP model's predictions (DialoGPT-small + pre-defined set of sentences).
\end{itemize} 

\begin{table}[!ht]
\small
\centering
\begin{tabular}{l|cc|cccc}
\toprule
\multicolumn{7}{c}{\textbf{Emotion-Push}} \\
\toprule
 & \textbf{\textsc{PPL}} & \textbf{\textsc{SES}} & \textbf{m-F1} & \textbf{m-NMC-F1} & \textbf{M-F1} & \textbf{M-NMC-F1}  \\ \hline
Baseline & 85.0 & 17.3 & 47.6 & 22.6 & 17.1 & 10.3 \\
 Oracle & \textbf{79.4} & \textbf{18.0} & \textbf{64.3} & \textbf{45.5} & \textbf{33.0} & \textbf{26.6} \\
 SACA & 80.3 & 16.6 & 49.8 & 21.6 & 15.7 & 8.4 \\
\toprule
\multicolumn{7}{c}{\textbf{DailyDialog}} \\
\toprule
  & \textbf{\textsc{PPL}} & \textbf{\textsc{SES}} & \textbf{m-F1} & \textbf{m-NMC-F1} & \textbf{M-F1} & \textbf{M-NM-F1C}  \\ \hline
 Baseline & 9.9 & 26.7 & 77.9 & 30.6 & 24.9 & 14.5 \\
 Oracle & \textbf{9.6} & 27.3 & \textbf{84.6} & \textbf{53.1} & \textbf{48.5} & \textbf{41.4} \\
 SACA & 9.7 & \textbf{27.6} & 77.9 & 30.8 & 26.0 & 15.8 \\
\toprule
\end{tabular}
\caption{Baseline + Oracle + Sentiment Aware Conversational Agent (SACA) results.}
\label{tab:tg-fullsystem-emotionpush}
\vspace{-2mm}
\end{table}

The results obtained can be observed in Table \ref{tab:tg-fullsystem-emotionpush}. The first conclusion we can take is that the appropriate sentiment, improves, indeed, scores in the generation process, as the \textit{Oracle} outperforms the \textit{Baseline}. Nevertheless, for both datasets, the introduction of the reply sentiment prediction is still far from the \textit{Oracle}. This proves the need for a better reply sentiment prediction model, which is key for the proper functioning of the proposed conversational agent. It should be noticed, however, that using the model with the pre-defined set of sentiment sentences in the \textit{SACA} still improves the \textit{perplexity} when compared to the \textit{Baseline}, which shows that it is the set of sentences that is being concatenated to the input that is improving the \textit{perplexity}, and not whether the model is receiving the correct sentiment or not. Regarding the \textit{sentence embedding similarity}, on the EmotionPush corpus it is noticeable that the \textit{SES} is better when the sentiment metrics and the \textit{perplexity} achieve better results. On the DailyDialog corpus this is not as perceptive, given the low fluctuation of the metric between the evaluated models. 

An example of the sentences generated by the developed sentiment-conditioned text generation model can be observed in Table \ref{tab:same-context-examples}. For the same context sentence we present the sentences generated by the model conditioned on each of the target sentiments. Additionally, we also show the sentence generated by the baseline, which is not conditioned on any sentiment. The model is capable of generating fluent and adequate answers for all sentiments. Furthermore, most sentences generated express the desired sentiment. Another interesting aspect is that the model only repeated keywords from the pre-defined set of sentiment sentences twice on a sentence generated conditioned on the \textit{fear} emotion. Interestingly, the \textit{fear} sentiment is the most underrepresented sentiment in the corpus, which might have made the model rely more on the provided set.

\begin{table}[ht!]
\centering
\begin{tabular}{ll}
\hline
\multicolumn{2}{|c|}{\textbf{Example 1}} \\ \hline
\textbf{Context} & How do you feel? \\ \toprule
\textbf{Baseline} & I feel great, thanks. \\ \hline
\textbf{Anger} & I'm sick of this city. \\ \hline
\textbf{Disgust} & I'm very nauseous. I've never felt so nauseous. \\ \hline
\textbf{Fear} & I'm \textcolor{red}{scared}. I'm \textcolor{red}{scared} of the future. \\ \hline
\textbf{Joy} & I like it very much. \\ \hline
\textbf{Sadness} & I'm really sorry, I'm really not feeling well. \\ \hline
\textbf{Surprise} & Oh, my god! My eyes are burning! \\ \hline

\end{tabular}
\caption{Example of sentences generated by the sentiment-conditioned text generation model. Words in \textcolor{red}{red} are present in the pre-defined set of sentences.}
\label{tab:same-context-examples}
\vspace{-2mm}
\end{table}

\subsection{Human Evaluation}

As mentioned in \cite{xie2021generating}, automatically evaluating empathetic conversational agents is a challenging task given the limitations of the automatic metrics. In particular, the most common text generation metrics evaluate the word/lexical overlap between the gold and generated sentences, and in a dialogue setting there can be many correct answers. Furthermore, the experiment previously reported on Table \ref{tab:same-context-examples}, further motivates the challenge of relying on these metrics to evaluate sentiment-aware conversational agents: feeding to the model the exact same context and conditioning it on different sentiments drastically changes the outcome of the generated sentences. For that matter, regarding the evaluation of our model, the most important aspects to evaluate were the adequacy given the previous context, and whether it expressed the desired sentiment. For this evaluation we were able to gather answers from seven annotators with a proficient English level. 

This evaluation was done by sampling at random 40 inputs from the test set of each corpus and retrieving the corresponding replies generated by four of the developed architectures: \textit{Baseline}, the DialoGPT-small model; \textit{Baseline Oracle}, the DialoGPT-small + \textit{tag} model; \textit{Oracle}, the DialogGPT-small + pre-defined set of sentences model; and \textit{SACA}, the sentiment-aware conversational agent.

In order to evaluate the adequacy of the reply we asked the annotators the following question: \say{Do the replies sound appropriate considering the context of the dialogue?}.
A similar process was followed to evaluate the sentiment of the sentences. Our goal with this evaluation was to assess if the model was able to generate sentences with the desired sentiment. To do so, we asked the annotators if the reply expressed the gold sentiment. The question asked was \textquotedblleft Does the reply represent the \texttt{<sentiment\_name>} emotion?\textquotedblright. Additionally, given the multitude of sentiments present that often can be interchanged, we asked the annotators to consider whether the sentence being evaluated could be said to express the asked sentiment. For example, \say{What?} could be used to express \textit{anger} or \textit{surprise}, depending on the tone used. We use a 2-point Likert scale for both questions (\textit{Yes} or \textit{No}).

\begin{table}[ht!]
\centering
\small
\begin{tabular}{l|cc|cc}
    \toprule
    & \multicolumn{2}{c}{\textit{EmotionPush}} & \multicolumn{2}{c}{\textit{DailyDialog}} \\
    \textbf{Model} & \textbf{Adequacy} & \textbf{Sentiment} & \textbf{Adequacy} & \textbf{Sentiment}\\ \hline
    \textit{Baseline} & 0.4292 & 0.325 & 0.4958 & 0.3708 \\
    \textit{Baseline Oracle} & 0.5042 & 0.5167 & \textbf{0.6542} & 0.7084 \\
    \textit{Oracle} & \textbf{0.6167} & \textbf{0.6583} & 0.6292 & \textbf{0.7625} \\
    \textit{SACA} & 0.3917 & 0.3542 & 0.5667 & 0.3292 \\
\toprule
\end{tabular}
\caption{Average EmotionPush and DailyDialog Scores.}
\label{tab:emotionpush-scores}
\vspace{-2mm}
\end{table}

The results obtained can be observed in Table \ref{tab:emotionpush-scores}. Given that we are using a Likert scale of 2-points, the reported scores correspond to the ratio of positive answers. E.g., an adequacy score of 0.6 for the \textit{Baseline} model means that 60\% of the generated sentences by this model were considered adequate. Similarly, sentiment score also corresponds to the ratio of positive answers. It is clear that the model that achieves the best performance was the \textit{Oracle}. In particular, this model improved the adequacy of the replies when compared to all other models. It also achieves the highest sentiment scores. It is also interesting to observe that, despite the accumulated error of the \textit{SACA} due to the reply sentiment prediction model, this model still outperforms the \textit{Baseline} on the sentiment metric. There also seems to exist a correlation between the sentiment metric and how adequate the replies are. The models that achieved a higher sentiment metric, also tend to be more adequate. Regarding the DailyDialog corpus, both \textit{Baseline Oracle} and \textit{Oracle} perform similarly. The improvements of both models when compared to the \textit{Baseline} are considerable across all metrics. 

\begin{table}[ht!]
\centering
\small
\begin{tabular}{lcccc}
    \toprule
    \textbf{} & \textbf{\textsc{PPL}} & \textbf{\textsc{SES}} & \textbf{m-NMC} & \textbf{\textsc{M-NMC}} \\ \hline
    \multicolumn{5}{c}{\textit{EmotionPush}} \\ \hline
    Adequacy & 0.18 & 0.9524 & 0.8015 & 0.8522 \\
    Sentiments & 0.01 & 0.9519 & 0.7992 & 0.8328 \\ \hline
    \multicolumn{5}{c}{\textit{DailyDialog}} \\ \hline
    Adequacy & -0.981 & 0.7992 & 0.8852 & 0.8589 \\
    Sentiments & -0.7345 & 0.4475 & 0.9966 & 0.9839 \\
\toprule
\end{tabular}
\caption{Pearson correlation between the automatic metrics and the human evaluation metrics.}
\label{tab:emotionpush-correlation}
\vspace{-6mm}
\end{table}

Although a correlation using four points (corresponding to pairs of the automatic and human metrics obtained for each evaluated model) is not ideal, and might not lead to statistically significant values, we finalize this analysis with the correlation between the obtained automatic metrics (perplexity, sentence embedding similarity, micro/macro-F1 without the majority class), and the human evaluation (adequacy and sentiment) by using the Pearson correlation to measure the linear relationship between them. We can observe the results in Table \ref{tab:emotionpush-correlation}.
The correlation between the sentiment automatic metrics and the human evaluation metrics is generally very high for both corpora. It is relevant to mention that on the EmotionPush corpus, the correlation between the human evaluation metrics and the perplexity is very low, while on the DailyDialog corpus is close to perfect. This showcases the difference in the quality of the text in both datasets. 
Furthermore, we highlighted in Section \ref{section:exp-text-generation} that models fine-tuned with formal and fluent data, such as the DailyDialog corpus, seem to perform well with simpler sentiment-conditioning approaches. In contrast, models fine-tuned with informal data, such as the EmotionPush corpus, seem to rely more on full sentiment sentences in order to perform well. This hypothesis is further validated with these experiments, since the annotators gave higher scores to the \textit{Oracle} model fine-tuned for the EmotionPush corpus, while the annotations gathered for the DailyDialog corpus showed similar results for the \textit{Oracle} and \textit{Baseline Oracle} approaches. 

The high correlation obtained in the human and automatic metrics highlights the importance of having a sentiment classification model as a metric to guide the development of the models.
\section{Related Work}\label{chap:related-work}


\subsection{Sentiment Classification} \label{section:rw-sentiment-classification}

The current state-of-the-art for sentiment classification is to use dense contextual word embedding models, such as BERT \cite{devlin2019bert}, via transfer learning. Other works on sentiment classification made use of context, speakers, speech acts and topics information \cite{kim2020contextual}, or further pre-trained BERT with a dataset similar to the target sentiment-labelled dataset \cite{huang2019emotionx}.

\subsection{Reply Sentiment Prediction} \label{section:rw-sentiment-prediction}

One critical aspect of systems that deal with sentiment-aware text generation in a dialogue context is how to define the appropriate sentiment for the upcoming reply. 
Several works approach this task (\textit{e.g.} \cite{herzig2016classifying} or \cite{bothe2017dialogue}) by adding additional dialogue and textual features, such as the time between interactions, or the emotion of previous sentences, to aid the model. To simulate the emotion transition in humans, the work described in \cite{wen2021automatically} makes use of the Valence-Arousal-Dominance emotion space, which encodes the emotion of words in a 3-dimensional vector space, to calculate the \say{emotion transition as the variation between the preceding emotion and the response emotion}. Predicting the next sentiment can also be part of the text generation model, as described in Section \ref{section:rw-text-generation}.


\subsection{Sentiment-Aware Text Generation} \label{section:rw-text-generation}

The current state-of-the-art in dialogue text generation consists in data-driven end-to-end models, which are capable of generating fluent, appropriate, and meaningful responses in a dialogue setting, by using previous context sentences as input to text generation models. One of the first successful approaches was proposed in \cite{vinyals2015neural}, which leverages the Seq2Seq architecture to predict an upcoming sentence. The work proposed in \cite{wolf2019transfertransfo} applies this idea to the Transformer architecture, and fine-tunes the GPT-2 model using two additional special tokens that are used to separate the sentences belonging to different speakers in the model's input. 

Regarding text generation with sentiment, the authors of \cite{rashkin2019towards} develop the EmpatheticDialogues dataset, fine-tune a GPT-2 model with it, and conclude that using a large-scale empathetic corpus enables the models to express appropriate emotions implicitly.
Alternative recent approaches make use of mechanisms that condition the text generation model on a desired sentiment. Some use a pre-defined set of rules or heuristics \cite{asghar2018affective, huang2018automatic}; others, data-driven approaches \cite{lubis2018eliciting, rashkin2018know, xie2021generating}. In \cite{asghar2018affective} it is proposed a conversational model which embeds words using the Valence-Arousal-Dominance emotion space \cite{mehrabian1996pleasure}, and explores decoding 
techniques that encourage diversity in candidate outputs. Furthermore, it designed a \say{training loss to explicitly train an affect-aware conversation model}, following three heuristics: minimizing affective dissonance (generated emotion should be similar to the input's emotion); maximizing affective dissonance (generated emotion should not be aligned with the input's emotion); and maximizing affective content (generated emotion should avoid being neutral). In \cite{huang2018automatic} it is adopted an \say{emotion mining from text} classifier, developed in \cite{yadollahi2017current}, to classify the emotions expressed in previous conversation context, and used this information, together with pre-defined mapping rules defined by the authors, to decide which emotion should be expressed. 
However, in \cite{xie2021generating} it is argued that approaches that rely on a pre-defined set of rules or heuristics, such as the previously mentioned, are not supported by psychology literature, and therefore emotional interactions in human-human conversations should be explored by using data-driven language models with large-scale emotional corpora. Some works leverage a multi-task approach that jointly trains an emotion encoder and a text generation model, which is conditioned on the emotional state assessed by the emotion encoder \cite{lubis2018eliciting, rashkin2018know}. Similarly to our work, \cite{xie2021generating} incorporates an intent predictor, which is separately trained from the text generation model, with the goal of deciding the intent for the reply to be generated. That intent is predicted based on previous context, and is then encoded and fed to the text generation model. Our work mainly differs by having a different model architecture, and exploring smaller datasets. Interestingly, their analysis on the intent predictor reaches similar conclusions to ours, which further shows that the answer to achieve a better performance in the reply sentiment prediction task might not be directly related to the amount of data used to train the models, and other methods should be explored.

\vspace{-2mm}

\section{Conclusion and Future Work}
\label{section:conclusion}

In this work we built a sentiment-aware conversational agent. In particular, we observed that using multiple context sentences on the input of the reply sentiment prediction model, and using a pre-defined set of sentiment sentences to condition the text generation model, improved performance when compared to baseline models. Furthermore, for text generation, we showed how our approach resulted in clear gains. Additionally, we observed how the reply sentiment prediction model is the bottleneck of the agent. Finally, we also performed a human evaluation on the developed models which corroborated the results obtained on the automatic evaluation, highlighting the necessity of metrics that evaluate the sentiment expressed during development.

As future work, we consider that improving the reply sentiment prediction model is crucial for a better performance of the conversational agent. To do so, we propose to explore different ways to incorporate the retrieval augmentation methods \cite{wang2021improving}, prompt-based learning \cite{han2021ptr}, or hybrid approaches with both data-driven and pre-defined rules.

\section*{Acknowledgements}

This work was supported by national funds through Fundação para a Ciência e a Tecnologia (FCT), project UIDB/50021/2020 (INESC-ID multi-annual funding); by FEDER, Programa Operacional Regional de Lisboa, Agência Nacional de Inovação (ANI), and CMU Portugal, under the project Ref. 045909 (MAIA).

\bibliographystyle{ACM-Reference-Format}
\bibliography{bibliography}

\end{document}